\documentclass{article}

\usepackage{microtype}
\usepackage{graphicx}
\usepackage{subfigure}
\usepackage{booktabs} 

\usepackage{hyperref}

\usepackage[accepted]{icml2023}

\usepackage{amsmath}
\usepackage{amssymb}
\usepackage{mathtools}
\usepackage{amsthm}
\usepackage{enumitem}
\usepackage[capitalize,noabbrev]{cleveref}

\theoremstyle{plain}
\newtheorem{theorem}{Theorem}[section]
\newtheorem{proposition}[theorem]{Proposition}

\theoremstyle{definition}

\theoremstyle{remark}

\usepackage[textsize=tiny]{todonotes}

\usepackage{amsmath,amsfonts,bm}

\def\eqref#1{equation~\ref{#1}}

\def\1{\bm{1}}

\def\rvw{{\mathbf{w}}}
\def\rvx{{\mathbf{x}}}

\DeclareMathAlphabet{\mathsfit}{\encodingdefault}{\sfdefault}{m}{sl}
\SetMathAlphabet{\mathsfit}{bold}{\encodingdefault}{\sfdefault}{bx}{n}

\def\gF{{\mathcal{F}}}
\def\gG{{\mathcal{G}}}

\def\gK{{\mathcal{K}}}
\def\gL{{\mathcal{L}}}

\def\gN{{\mathcal{N}}}

\def\gP{{\mathcal{P}}}
\def\gQ{{\mathcal{Q}}}

\def\gT{{\mathcal{T}}}

\def\gW{{\mathcal{W}}}

\def\sC{{\mathbb{C}}}

\def\sR{{\mathbb{R}}}

\newcommand{\E}{\mathbb{E}}

\newcommand{\df}{\mathrm{d}}
\newcommand{\din}{d_{\mathrm{in}}}

\theoremstyle{plain}

\icmltitlerunning{Fast Sampling of Diffusion Models via Operator Learning}

\begin{document}

\twocolumn[
\icmltitle{Fast Sampling of Diffusion Models via Operator Learning}




\begin{icmlauthorlist}
\icmlauthor{Hongkai Zheng}{yyy}
\icmlauthor{Weilie Nie}{comp}
\icmlauthor{Arash Vahdat}{comp}
\icmlauthor{Kamyar Azizzadenesheli}{comp}
\icmlauthor{Anima Anandkumar}{yyy,comp}
\end{icmlauthorlist}

\icmlaffiliation{yyy}{Caltech}
\icmlaffiliation{comp}{NVIDIA}

\icmlcorrespondingauthor{Hongkai Zheng}{hzzheng@caltech.edu}

\icmlkeywords{Machine Learning, ICML}

\vskip 0.3in
]



\printAffiliationsAndNotice{}  

\begin{abstract}
Diffusion models have found widespread adoption in various areas. However, their sampling process is slow because it requires hundreds to thousands of network evaluations to emulate a continuous process defined by differential equations. In this work, we use neural operators, an efficient method to solve the probability flow differential equations, to accelerate the sampling process of diffusion models. Compared to other fast sampling methods that have a sequential nature, we are the first to propose a parallel decoding method that generates images with only one model forward pass. We propose \textit{diffusion model sampling with neural operator} (DSNO) that maps the initial condition, i.e., Gaussian distribution, to the continuous-time solution trajectory of the reverse diffusion process. To model the temporal correlations along the trajectory, we introduce temporal convolution layers that are parameterized in the Fourier space into the given diffusion model backbone. We show our method achieves state-of-the-art FID of 3.78 for CIFAR-10 and 7.83 for ImageNet-64 in the one-model-evaluation setting. 
\end{abstract}

\section{Introduction}

\begin{figure*}[t]
        \centering
        \includegraphics[width=0.92\textwidth]{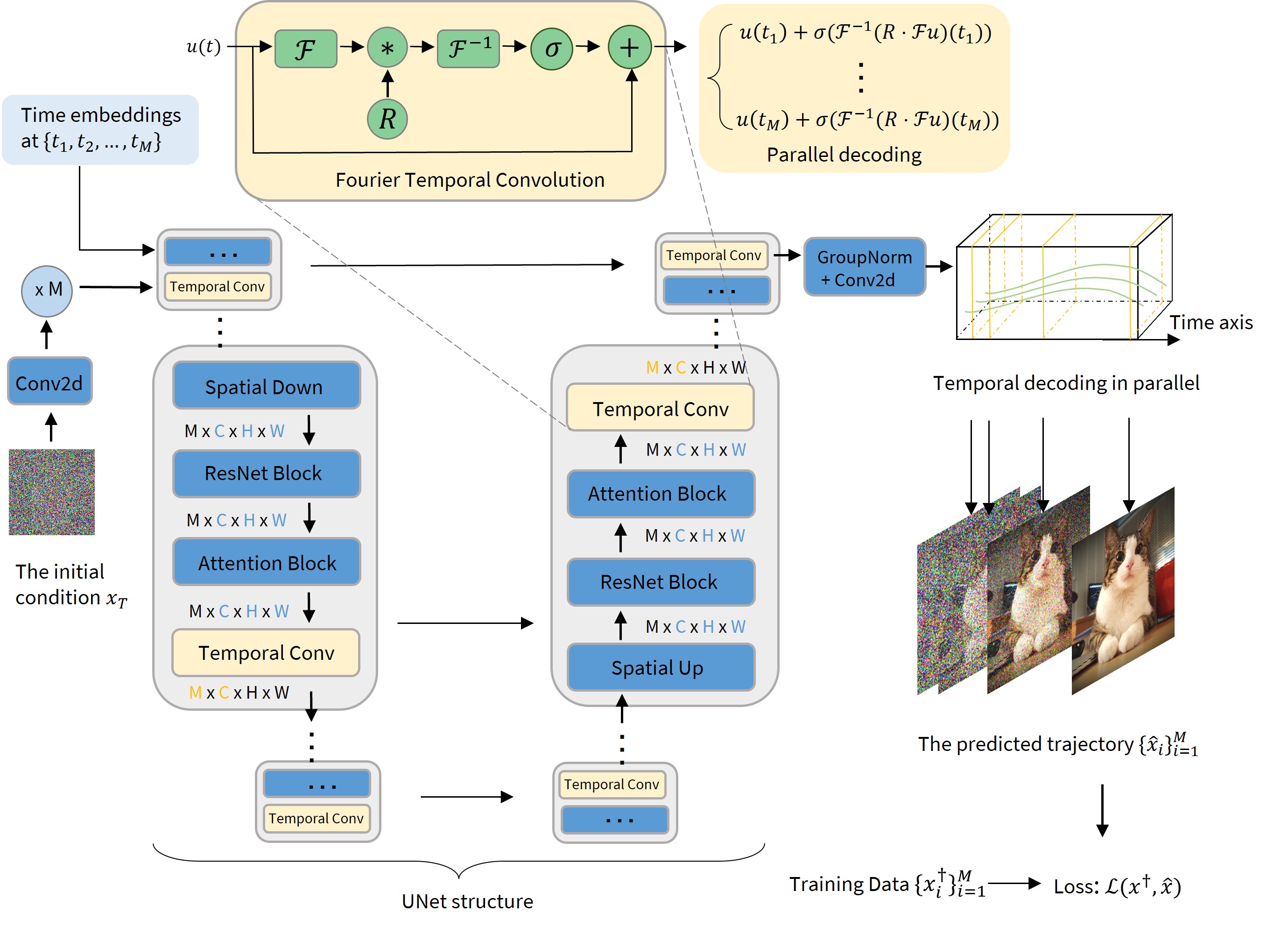}
        \caption{Illustration of the architecture and training pipeline of DSNO. The architecture of DSNO is built on top of any existing diffusion model architecture, where blue blocks are from the existing diffusion U-Net backbone and yellow blocks are the proposed temporal convolution layers.
        Suppose the temporal domain is discretized into $M$ points $\{t_1,\dots,t_M\}$,
        for each feature map, the temporal convolution layers operate on the temporal and channel dimensions (M$\times$C) and the other blocks operate on the pixel and channel dimensions (C$\times$H$\times$W). The symbols $\gF$ and $\gF^{-1}$ refer to the Fourier transform and inverse Fourier transform, respectively. 
        $R$ is a complex-valued parameter that represents a kernel function in Fourier space. For ease of notation, $x_i$ represents the solution at time $t_i$, that is $x(t_i)$. 
        Inside each temporal convolution layer, we apply the idea of parallel decoding: Given input function $u(t)$, the Fourier coefficients $R\cdot \gF u$ is the same for all $t_i, i=1,\dots, M$. Therefore, the temporal convolution layer can output the representations at different time locations in the trajectory in a single forward pass by evaluating the output function at queried points in parallel.}
        \label{fig:dfno_arch}
\end{figure*}

Diffusion models~\citep{sohl2015deep,ho2020denoising}, also known as score-based generative models~\citep{song2020score}, have emerged as a powerful generative modeling framework in various areas. They have achieved state-of-the-art (SOTA) performance in many applications including image generation \cite{dhariwal2021diffusion}, molecule generation \cite{xu2022geodiff}, audio synthesis \cite{kong2021diffwave} and model robustness~\cite{nie2022DiffPure}. However, sampling from diffusion models requires hundreds of neural network evaluations, making them slower by orders of magnitude compared to other generative models such as generative adversarial networks (GANs) \cite{goodfellow2020generative}. Accelerating sampling in diffusion models remains a challenging but important problem, especially when applying them to time-sensitive downstream applications such as AI for art and design \cite{ramesh2022hierarchical} or generative models for decision making \cite{ajay2022conditional}. 

Existing methods for fast sampling of diffusion models can be summarized into two main categories: 1) \emph{training-free sampling methods}~\citep{song2020denoising,lu2022dpm} and 2) \emph{training-based sampling methods}~\citep{luhman2021knowledge,salimans2021progressive,xiao2021tackling}. Specifically, the training-free methods focus on reducing the number of discretization steps from a numerical perspective while solving the stochastic differential equations (SDE) or probability flow ordinary differential equations (ODE). However, even the best well-designed numerical solvers \cite{lu2022dpm, karras2022elucidating} still need 10$\sim$30 model evaluations such that the approximation error is small enough for an acceptable sampling quality.
On the other hand, training-based methods train a  surrogate network to replace some parts of the numerical solver or even the whole solver. Particularly, progressive distillation \cite{salimans2021progressive} has made a big step towards real-time sampling (e.g., decent results with 4 steps) but it still has a sequential nature like conventional numerical solvers. 

The goal of this work is to develop a fast and parallel sampling method for diffusion models \emph{with only one model evaluation}. By parallel, we mean that our method can decode images at different time locations in the trajectory in parallel and hence, generate the final solution using only one model evaluation. The major challenge here arises from the difficulty of solving a complicated and large-scale differential equation, which typically requires many discrete time steps to emulate accurately from a numerical approximation perspective. 

In this paper, we employ the recent advances in neural operators for solving differential equations to overcome this challenge. 
Neural operators \citep{li2020neural,kovachki2021neural}, especially the Fourier neural operator (FNO)~\cite{li2020fourier} have shown several orders of magnitude speedup over conventional solvers. This class of models enables learning maps between spaces of functions and is shown to be discretization invariant, allowing them to work with different resolutions of data without changing the model parameters, and can approximate any given nonlinear continuous operator~\cite{kovachki2021universal}. 

The FNO allows for parallel decoding: i.e. the outputs at all locations of the trajectory can be simultaneously evaluated. This is a property that none of the previous sampling methods for diffusion models enjoy.   
In this work, we propose a neural operator for diffusion model sampling (DSNO) that maps the initial conditions (i.e. Gaussian distribution) to the solution trajectories and we show its effectiveness in both unconditional and class-conditional image generation.

 \paragraph{Our contributions.}
\begin{itemize}[leftmargin=*]
    \item We propose a neural operator for the fast sampling of diffusion models (DSNO) that can sample high-quality images with one model evaluation. 
    \item We introduce temporal convolution blocks parameterized in Fourier space, which can be easily combined with any existing neural architectures of diffusion models to build a neural operator backbone for DSNO. Furthermore, our proposed temporal convolution blocks are lightweight and only increase the model size by 10\%.
    \item For the first time, we propose a parallel decoding method to generate the trajectories of images using continuous function representation, which enables generation of the final solution in one model evaluation.
    \item Our proposed DSNO achieves new state-of-the-art FID scores of 3.78 for CIFAR-10 and 7.83 for ImageNet-64 in the setting of single-step-generation of diffusion models.
\end{itemize}

Finally, we note that DSNO leverages parallel decoding temporally to generate the solution trajectory by evaluating the output function at different time steps in parallel. This is in contrast to the prior training-based methods that have a sequential nature and predict the trajectory step by step.  
We believe that DSNO with parallel decoding is a key step for the real-time sampling of diffusion models, potentially benefiting many interactive applications. 


\section{Background}
\paragraph{Score-based generative models.} We consider the general class of score-based generative models in a unified continuous-time framework proposed by \citet{song2020score}, which includes different variants of diffusion models \cite{sohl2015deep, ho2020denoising}. In this paper, we will use the word score-based models interchangeably with diffusion models. Suppose the data distribution is $p_{\mathrm{data}}$. The forward pass is a diffusion process $\{\rvx(t)\}$ starting from $0$ to $T$ which can be expressed as
\begin{equation}
    \df \rvx = f(\rvx, t) \df t + g(t) \df \rvw_t,
\end{equation}
where $\rvw_t$ is the standard Wiener process, and $f(\cdot,t): \mathbb{R}^d \rightarrow \mathbb{R}^d$ and $g(\cdot): \mathbb{R}\rightarrow \mathbb{R}$ are the drift and diffusion coefficients respectively. Diffusion models choose $f$ and $g$ such that $\rvx(0)\sim p_{\mathrm{data}}$ and $\rvx(T) \sim \mathcal{N}(0, \textbf{I})$. 
 \citet{song2020score} show that the following probability flow ODE produces the same marginal distributions $p_t(\rvx)$ as that of the diffusion process:
\begin{equation}
    \df \rvx = f(\rvx, t) \df t  - \frac{1}{2}g(t)^2 \nabla_{\rvx}\log p_t(\rvx) \df t.
    \label{eq:prob_ode}
\end{equation}
The sampling process eventually becomes solving the probability flow ODE~\ref{eq:prob_ode} from $T$ to $0$ given the initial condition $\rvx(T)$. Furthermore, $f(\rvx, t)$ often has the affine form $f(\rvx,t) = h(t) \rvx$, where $h: \mathbb{R} \rightarrow \mathbb{R}$. We can simplify the equation~\ref{eq:prob_ode} into a semi-linear ODE. Integrating both sides over time gives the explicit form of solution for any $t<s$:
\begin{equation}
    \rvx(t) = \phi(t, s) \rvx(s) - \int_{s}^{t} \phi(t, \tau) \frac{g(\tau)^2}{2} \nabla_{\rvx}\log p_{\tau}(\rvx) \df \tau, 
    \label{eq:ode_soln}
\end{equation}
where $\phi(t, s) = \exp\left(\int_{s}^{t} h(\tau) \df \tau \right)$. The ODE can be solved using numerical solvers such as Euler's method, multistep methods, and Heun's 2\textsuperscript{nd} method. 
The score function $\nabla_{\rvx} \log p_t(\rvx) $ is usually parameterized by $
    \hat{\epsilon}_{\theta}(\rvx_t) \approx - \sigma_t \nabla_{\rvx} \log p_t(\rvx)$, where $\sigma_t$ is the noise schedule~\cite{song2020score, ho2020denoising}.

\vspace{-0.1in}
\paragraph{Fourier neural operator.}
Fourier neural operator \cite{li2020fourier} is one of the state-of-the-art data-driven methods for solving PDEs, which has shown great speedup over conventional PDE solvers in many scientific problems by learning a parametric map between two Banach spaces from data. They are constructed as a stack of kernel integration layers where the kernel function is parameterized by learnable weights. Let $D$ be a bounded domain, e.g., $[0,T]$ and $a: D\rightarrow \mathbb{R}^{\din}$ denote an input function. A Fourier neural operator $\gG_{\theta}$, parameterized with learnable parameters $\theta$, is an $L$ layered neural operator of the following form,
\begin{equation}
    \gG_{\theta} \coloneqq \gQ \circ \sigma(\gW_{L} + \gK_{L})\circ \dots \circ \sigma(\gW_{1} + \gK_{1}) \circ \gP,
    \label{eq:fno_layers}
\end{equation}
where the lifting operator $\gP$, projection operator $\gQ$, and residual connections $\gW_{i}, i\in \{1, \dots, L\}$ are pointwise operators parameterized with neural networks, and $\sigma$ is a fixed nonlinear activation function. $\gK_{i}$ is an integral kernel operator parameterized in Fourier space such that for a given $v_i$, an input function to the $i$'th layer, we have,
\begin{equation}
    (\gK v_i)(t) = \gF^{-1}\left(R_i \cdot (\gF v_i)\right)(t), \forall t\in D 
    \label{eq:fourier_conv}
\end{equation}
where $\gF$ and $\gF^{-1}$ are the Fourier transform and inverse Fourier transform on $D$, $R_i$ is a trainable parameter that parameterizes a kernel function in Fourier space. Given an input function $a$, we first apply the lifting point-wise operator $\gP$ that expands the co-dimension of the input function $a$, followed by $L$ layers of global integral operators accompanied with pointwise non-linearity operation $\sigma$. The result of the global integration layers is passed to the local and pointwise projection layer $\gQ$ to compute the output function. This architecture is shown to possess the crucial discretization invariance and universal approximation properties of universal operators~\citep{kovachki2021universal,kovachki2021neural}.

\section{Learning the trajectory with neural operator}
\paragraph{Problem statement.}
Our goal is to learn a neural operator that given any initial condition $\rvx(T)\sim \mathcal{N}(0,\mathbf{I})$, predicts the probability flow trajectory $\{\rvx(t)\}_{s}^{0}$ with time flowing from $s$ to $0$ defined in equation~\ref{eq:ode_soln}, where the endpoint $\rvx(0) \in \sR^d$ is the data. 
Let $D=[0, s], 0 < s \leq T$ be the temporal domain. 
Let $\mathcal{A}$ be the finite-dimensional space of the initial condition, and $\mathcal{U}=\mathcal{U}(D; \mathbb{R}^d)$ denote the space of the target continuous time functions with output value in $\sR^d$. 
We build a neural operator $\gG_{\theta}$ parameterized by $\theta$ to approximate the solution operator $\gG^\dagger$ by minimizing the error as follows
\begin{equation}
    \min_{\theta} \E_{\rvx_T \sim \gN(0,\mathbf{I})} \gL\left(\gG_{\theta}\left(\rvx_T\right) - \gG^\dagger\left(\rvx_T\right)\right),
    \label{eq:}
\end{equation}
where 
$\mathcal{L}: \mathcal{U}  \rightarrow \mathbb{R}_+$ 
is some loss functional such as $L^{p}$-norm for some $p\geq 1$. From the exact solution $\rvx(t)$ in equation~\ref{eq:ode_soln}, we know the solution operator $\gG^\dagger: \mathcal{A} \rightarrow \mathcal{U}$ exists and is a unique weighted integral operator of the score function. In other words, 
the solution operator corresponds to the underlying diffusion ODE, i.e., a mapping from a $\rvx(T)\sim \mathcal{N}(0,\mathbf{I})$ to the probability flow trajectory $\{\rvx(t)\}_{s}^{0}$. 
It is a regular operator, i.e., a member of operator set in the neural operator theory that can be approximated~\citep{kovachki2021neural,kovachki2021universal}. More formally,

\begin{proposition} [\citet{kovachki2021neural,kovachki2021universal}]
The class of neural operators defined in equation~\ref{eq:fno_layers} approximates the solution map of the diffusion ODE, i.e., a mapping from $\rvx(T)\sim \mathcal{N}(0,\mathbf{I})$ to the probability flow trajectory $\{\rvx(t)\}_{s}^{0}$, arbitrarily well. 
\end{proposition}
This implies that the proposed architecture has the required capacity to learn to output the continuous time probability flow trajectory $\{\rvx(t)\}_{s}^{0}$ in one model call.



\vspace{-0.1in}
\paragraph{Temporal convolution block in Fourier space.}
Inspired by the weighted integral form of the exact ODE solution in equation~\ref{eq:ode_soln}, we build our temporal convolution block with Fourier integral operator $\mathcal{K}$ to efficiently model the trajectory.  
Given an input function $u: D\rightarrow \sR^d$, our temporal convolution layer $\gT$ is defined as
\begin{equation}
    (\gT u)(t) = u(t) +  \sigma \left(\left(\gK u \right)(t)\right),
\end{equation}
where $\sigma$ is a point-wise nonlinear function, and $\gK$ is a Fourier convolution operator defined in equation \ref{eq:fourier_conv} parameterized by $R$. Note that our proposed temporal convolution layer differs slightly from the FNO layer given in equation \ref{eq:fno_layers}. 
Specifically, we move the nonlinear activation function  right after the Fourier convolution operator $\gK$ and replace the linear pointwise operator $\gW$ with an identity shortcut, which preserves the high-frequency information without extra cost and also leads to a better optimization landscape~\cite{he2016deep}. 
We have not observed the advantages of using a more general linear layer. The identity map is shown to be sufficient and more attractive because it is computationally efficient. Furthermore, we note that, by convolution theorem, we have 
\begin{equation}
   (\gK u)(t) = \int_{D} (\mathcal{F}^{-1}R)(\tau)u(t-\tau) \df \tau, \forall t\in D. \label{eq:fourier_int}
\end{equation}
Notably, the integral form in equation~\ref{eq:fourier_int} inherently possesses a structural similarity to the core diffusion process in equation~\ref{eq:ode_soln}, meaning that the temporal convolution layer implicitly parameterize the ODE solution trajectory. 

In practice, we use the discrete Fourier transforms for computational efficiency.
Suppose the temporal domain $D$ is discretized into $M$ points. 
For ease of understanding, we also assume the codomains of the input and output functions of the temporal convolution block are both in $\sR^d$. 
 The input function $u(t)$ is represented as a tensor in $\sR^{M\times d}$. $R$ is a complex-valued parameter in $\sC^{J\times d \times d}$, where $J$ is the maximal number of modes that we can choose. For all $u$, we truncate the modes higher than $J$ and then have $\gF(u) \in \sC^{J\times d}$. The pointwise product of Fourier transforms of input and kernel functions is given by
\begin{align}
    R \cdot (\gF u)_{j,k} = \sum_{l=1}^{d} R_{j,k,l} (\gF u)_{j,l}, 
\end{align}
for all $ j\in \{1,\dots, J\}, k\in \{1,\dots, d\}$. Accordingly, $\gF$ and $\gF^{-1}$ are realized by the fast Fourier transform algorithm. 
Figure~\ref{fig:dfno_arch} demonstrates the implementation details of the temporal convolution layers. Note that the temporal convolution layer only operates over the temporal dimension and hidden feature channel dimension and thus treats the pixel dimension as the same as the batch dimension. In other words, $d$ in the above example corresponds to the number of channel dimensions in practice.

\vspace{-0.1in}
\paragraph{Architecture of DSNO.}
As demonstrated in Figure~\ref{fig:dfno_arch}, the architecture of DSNO is built on top of any existing architecture of diffusion models, by adding our proposed temporal convolution layers to each level of the U-Net structure. The dark blue blocks are the modules in the existing diffusion model backbone, which treat the temporal dimension the same as the batch dimension and only work on the pixel and channel dimension. The yellow blocks are the Fourier temporal convolution blocks, which only perform on the temporal and channel dimension. Therefore, our model is highly parallelizable and adds minimal computation complexity to the original backbone. Again, suppose the temporal domain is discretized into $\{t_1, \dots, t_M\}$. The DSNO takes as input the time embeddings at these times and the initial condition. The feature map of the first convolution layer is repeated $M$ times over the temporal dimension as the initial feature at different times.
Each feature representation is combined with the corresponding time embedding in the following ResNet blocks. 

\vspace{-0.1in}
\paragraph{Training of DSNO}
Training DSNO is a standard operator learning setting. The training objective is a weighted integral of the error: 
\begin{equation}
    \min_{\theta} \E_{\rvx_T\sim \gN(0, \mathbf{I})} \int_D \lambda(t)\|\gG_{\theta}(\rvx_T)(t) - \gG^{\dagger}(\rvx_T)(t)\| \df t, 
\end{equation}
where $\theta$ is the parameter of DSNO, $\lambda(t)$ is the weighting function, $\rvx_{T}$ is the initial condition, and $\|\cdot\|$ is a norm. In practice, we optimize over $\theta$ to minimize the empirical-risk similar to ~\citet{kovachki2021neural}:
\begin{equation}
    \min_{\theta} \frac{1}{N}\sum_{j=1}^{N}\frac{1}{M}\sum_{i=1}^{M} \lambda(t_i)\|\gG_{\theta}(\rvx_T^{(j)})(t_i) - \gG^{\dagger}(\rvx_T^{(j)})(t_i)\|,
\end{equation}
where $\{t_1,\dots, t_M\}$ are discrete points in the temporal domain, and $\gG^{\dagger}(\rvx_T^{(j)})(t_i)$ can be generated from any existing solver or sampling method. 
\vspace{-0.1in}
\paragraph{Parallel decoding}
As shown in the top two yellow blocks in Figure~\ref{fig:dfno_arch}, the proposed Fourier temporal convolution block can predict images at different times in parallel. 
Given any input function $u(t)$, we can compute the Fourier coefficient $R\cdot \gF u$ and then call the inverse Fourier transform at all $t_i$ in parallel to generate output for different times at once. 
Plus, the other modules of DSNO treat temporal dimension like batch dimension and can perform in parallel for different $t_i$s. Therefore, DSNO is capable of efficient parallel decoding. 
Note that the effectiveness of our parallel decoding is based on the fact that the solutions of the diffusion ODE at different times are conditionally independent given the initial condition. Parallel decoding has shown its efficiency in transformers-based models~\cite{chang2023muse} and language models~\cite{ghazvininejad2019mask} for discrete tokens generation in the spatial domain. DSNO is the first parallel decoding method for continuous diffusion ODE trajectory, which is in temporal domain.   

\vspace{-0.1in}
\paragraph{Compact power spectrum.}
We examine the spectrum of the probability flow ODE trajectories generated from several publicly available pre-trained diffusion models in the literature, and observe that the ODE trajectories always have a compact energy spectrum over the temporal dimension. See more details in Appendix~\ref{sec:spectrum}. 
The smoothness of the diffusion ODE trajectory means the high-frequency modes do not contribute much to the learning objective. Therefore, DSNO built upon the stacks of Fourier temporal convolution layers can model the underlying solution operator of diffusion ODEs more efficiently with a relatively small number of discretization steps $M$. 




\begin{table}[t]
\caption{Comparison of fast sampling methods on CIFAR-10 for diffusion models in the literature. The FID score is computed with the original FID implementation to compare with the other methods. NFE: number of function evaluations.}
\label{tb:fid_cifar}
\vskip 0.1in
\begin{center}
\begin{small}
\begin{tabular}{lccc}
\toprule
Method & NFE & FID & Model size \\
\midrule
Ours    & 1 &  \textbf{3.78} & 65.8M   \\ 
\midrule
Knowledge distillation\\\cite{luhman2021knowledge}  & 1 & 9.36 & 35.7M\\
\midrule
Progressive distillation\\\cite{salimans2021progressive}    & 1 & 9.12 & 60.0M  \\
        & 2 & 4.51  & \\
        & 4 & 3.00  & \\ 
\midrule 
LSGM~\cite{vahdat2021score} & 147 & 2.10  & 475.0M\\
\midrule
GGDM + PRED + TIME\\ \cite{watson2021learning}                      & 5                 & 13.77   &  35.7M  \\
                                         & 10                 & 8.23    &   \\ 
\midrule
DDIM~\cite{song2020denoising}   & 10                & 13.36   &  35.7M \\ 
                                & 20           & 6.84   &   \\
                                & 50 & 4.67 & \\
\midrule 
SN-DDIM~\cite{bao2022estimating} & 10 & 12.19 & 52.6M \\ 
\midrule 
FastDPM~\cite{kong2021fast} & 10    & 9.90 & 35.7M \\ 
\midrule
DPM-solver~\cite{lu2022dpm} & 10 & 4.70 & 35.7M\\ 
\midrule 
DEIS~\cite{zhang2022fast} & 10 & 4.17 & - \\
\toprule 
\textbf{Diffusion + GAN} \\
\midrule
TDPM~\cite{zheng2022truncated} & 5 & 3.34 & 35.7M\\
DDGAN~\cite{xiao2021tackling} & 4 & 3.75 & - \\
\bottomrule
\end{tabular}
\end{small}
\end{center}
\vskip -0.05in
\end{table}
\section{Experiments}
In our experiments, we examine the proposed method on both unconditional and conditional image generation tasks.
We show that our method dramatically accelerates the sampling process of diffusion models, compared to existing fast sampling methods including both training-free and training-based approaches. Our code is available at \href{https://github.com/devzhk/DSNO-pytorch}{https://github.com/devzhk/DSNO-pytorch}. 

\begin{table*}[t]
\caption{Comparison of fast sampling methods on class-conditional ImageNet-64 for diffusion models in the literature. The results of DDIM and EDM are reported by \citet{karras2022elucidating} using the pre-trained model \cite{dhariwal2021diffusion}.}
\label{tb:fid_imagenet}
\vskip 0.1in
\begin{center}
\begin{small}
\begin{tabular}{lcccc}
\toprule
Method & Model evaluations & FID score & Recall & Model size\\
\midrule
Ours    & 1 &  \textbf{7.83} & \textbf{0.61} & 329.2M\\
\midrule
Progressive distillation~\cite{salimans2021progressive}    & 1 & 15.99 & 0.60 &  295.9M\\
        & 2 & 7.11 & 0.63 & \\
        & 4 & 3.84 &  0.63 & \\ 
\midrule
EDM~\cite{karras2022elucidating}                   &  79                 & 2.44  & 0.67   &  295.9M\\
\midrule
DDIM~\cite{song2020denoising}   & 32                & 5.00   & -  & 295.9M \\
\midrule
BigGAN-deep~\cite{brock2018large} & 1 & 4.06 & 0.48 & \\
ADM~\cite{dhariwal2021diffusion} & 250 & 2.07 & 0.63 & 295.9M \\ 
\bottomrule
\end{tabular}
\end{small}
\end{center}
\vskip -0.1in
\end{table*}

\begin{table}[t]
\caption{One model evaluation cost tested on V100. We compare the time cost of a single forward pass of DSNO and the corresponding original backbone. The reported results are averaged over 20 runs. The baseline models are from~\citet{salimans2021progressive}.}
\label{tb:speed_cifar}
\begin{center}
\begin{small}
\begin{tabular}{lccc}
\toprule
Backbone & Runtime & Model size \\
\midrule
CIFAR-10   & 0.033s & 60.00M \\
 DSNO-CIFAR-10 (ours)   & 0.050s & 65.77M\\
 \midrule
ImageNet64   & 0.066s & 295.90M  \\
DSNO-ImageNet-64 (ours) & 0.080s &  329.23M\\
\bottomrule
\end{tabular}
\end{small}
\end{center}
\vskip -0.1in
\end{table}

\subsection{Experimental setup}
We first randomly sample a training set of ODE trajectories using the pre-trained diffusion model to be distilled. 
We then build the network backbone for DSNO by simply adding the proposed temporal convolution layers to the above diffusion model. We initialize the modules from the existing architecture  with the pre-trained weights. 
As for the activation function in the temporal convolution layer, we use the leaky rectified linear unit (LeakyReLU) for $\sigma$.
We mainly use $\ell^1$ loss for the experiments on CIFAR10 and ImageNet-64. We also experiment with LPIPS~\cite{zhang2018unreasonable} loss on CIFAR10 like one concurrent work~\cite{song2023consistency} does. Regarding the choice of the loss weighting function, we set $\lambda(t) = \frac{\alpha_t}{\sigma_t}$, which is the square root of the SNR loss weighting used in the original diffusion model~\cite{salimans2021progressive}. We take the square root because our loss function is not squared. 
We use a batch size of 256 for CIFAR-10 experiments, a batch size of 2048 for ImageNet experiments, and a batch size of 128 by default in our ablation study. We use the same base learning rate of 0.0002, learning rate warmup schedule, and $\beta_1,\beta_2$ of Adam~\cite{kingma2014adam} as used in the diffusion model training without tuning these hyperparameters. 
\vspace{-0.1in}
\paragraph{Evaluation metric}
We use the Frechet inception distance (FID) \cite{heusel2017gans} to evaluate the quality of generated images. FID score is computed by comparing 50,000 generated images against the corresponding reference statistics of the dataset.  We use the ADM’s TensorFlow evaluation suite~\citep{dhariwal2021diffusion} and EDM's evaluation code~\cite{karras2022elucidating} to compute FID-50K with the same reference statistics. We also report Recall~\cite{kynkaanniemi2019improved} as the secondary metric of mode coverage for the experiments on ImageNet-64.

\subsection{Unconditional generation: CIFAR-10}

\paragraph{Trajectory data collection.} We first generate 1 million trajectories with 512-step DDIM \cite{song2020denoising} using the pre-trained diffusion model proposed by~\citet{salimans2021progressive}, and use it to train DSNO. The FID score of the training set is 2.51, computed over the first 50k data points in the training set. 

\vspace{-0.1in}
\paragraph{Sampling quality and speed.}
Table~\ref{tb:fid_cifar} compares the proposed DSNO trained with a temporal resolution of 4 against both training-based and training-free sampling methods in terms of FID and the corresponding number of model evaluations. DSNO clearly outperforms all the baselines with only one model evaluation and even achieves a better FID score than 2-step progressive distillation models. Furthermore, we compare the cost of one single forward pass of both DSNO and the original backbone\footnote{The progressive distillation only has JAX implementation. We implement its backbone in Pytorch and port the pre-trained weights from the official JAX checkpoint so that we can make a fair speed comparison within the same framework. } on a V100 in a standard AWS p3.2xlarge instance. For the speed test, we do 20 warm-up runs to avoid the potential inconsistency arising from the built-in cudnn autotuner. Since the time cost of progressive distillation grows linearly with the number of sampling steps, we can easily calculate the speedup of DSNO over the progressive distillation from Table~\ref{tb:speed_cifar}. DSNO is 2.6 times faster than the 4-step progressive distillation model and 1.3 times faster than 2-step progressive distillation model. Compared to hybrid models that combine GAN and diffusion models, DSNO achieves comparable performance with at most one-fourth number of model evaluations. 



\begin{figure}[t]
        \centering
        \includegraphics[width=0.97\columnwidth]{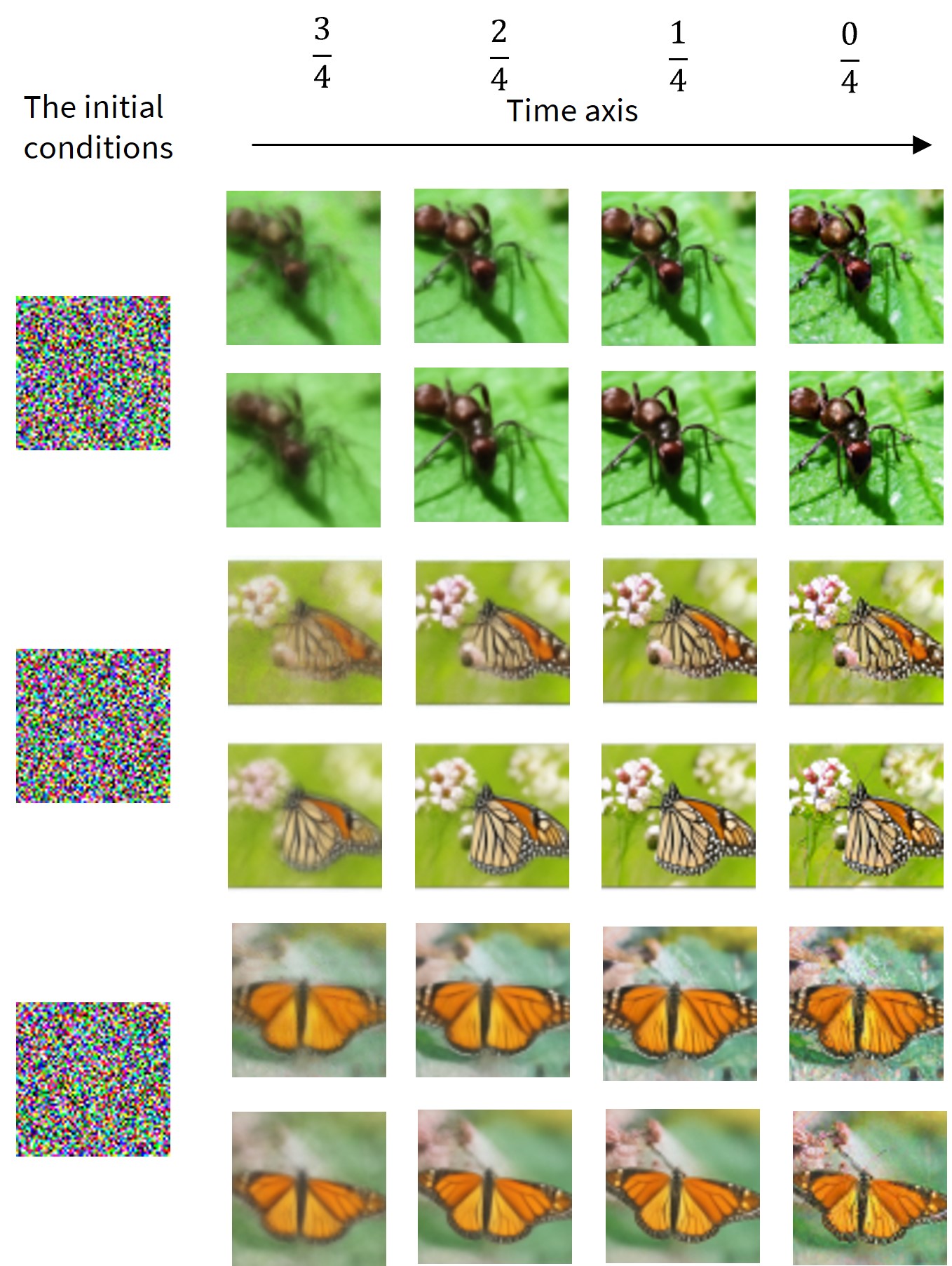}
        \vspace{-3pt}
        \caption{Comparison between the trajectory predicted by DSNO and that from the original solver on ImageNet-64, for the fixed random seed with a temporal resolution 4. Upper row: the prediction by DSNO. Lower row: the trajectory generated by solver. }
        \label{fig:demo_imagenet_traj}
        \vspace{-5pt}
\end{figure}

\subsection{Conditional generation: ImageNet-64}
\paragraph{Trajectory data collection.} We generate 2.3 million trajectories with 16-step progressive distillation \cite{salimans2021progressive} using the pre-trained diffusion model from its official code base. The FID score of the generated training set is 2.70, computed over the first 50k training data points. 
\vspace{-0.1in}
\paragraph{Sampling quality and speed.} Table~\ref{tb:fid_imagenet} compares DSNO trained with a temporal resolution of 4 against the recent advanced fast sampling methods for diffusion models. 
DSNO clearly outperforms 1-step progressive distillation model and archives comparable FID  2-step models of progressive distillation with only one model evaluation. From Table~\ref{tb:speed_cifar}, DSNO has 1.7 times speedup over progressive distillation. The recall of DSNO is comparable to ADM's, showing that DSNO inherits the original diffusion model's diversity/mode coverage as it learns to solve the probability flow ODE. 

\begin{figure}[ht]
        \centering
        \includegraphics[width=0.8\columnwidth]{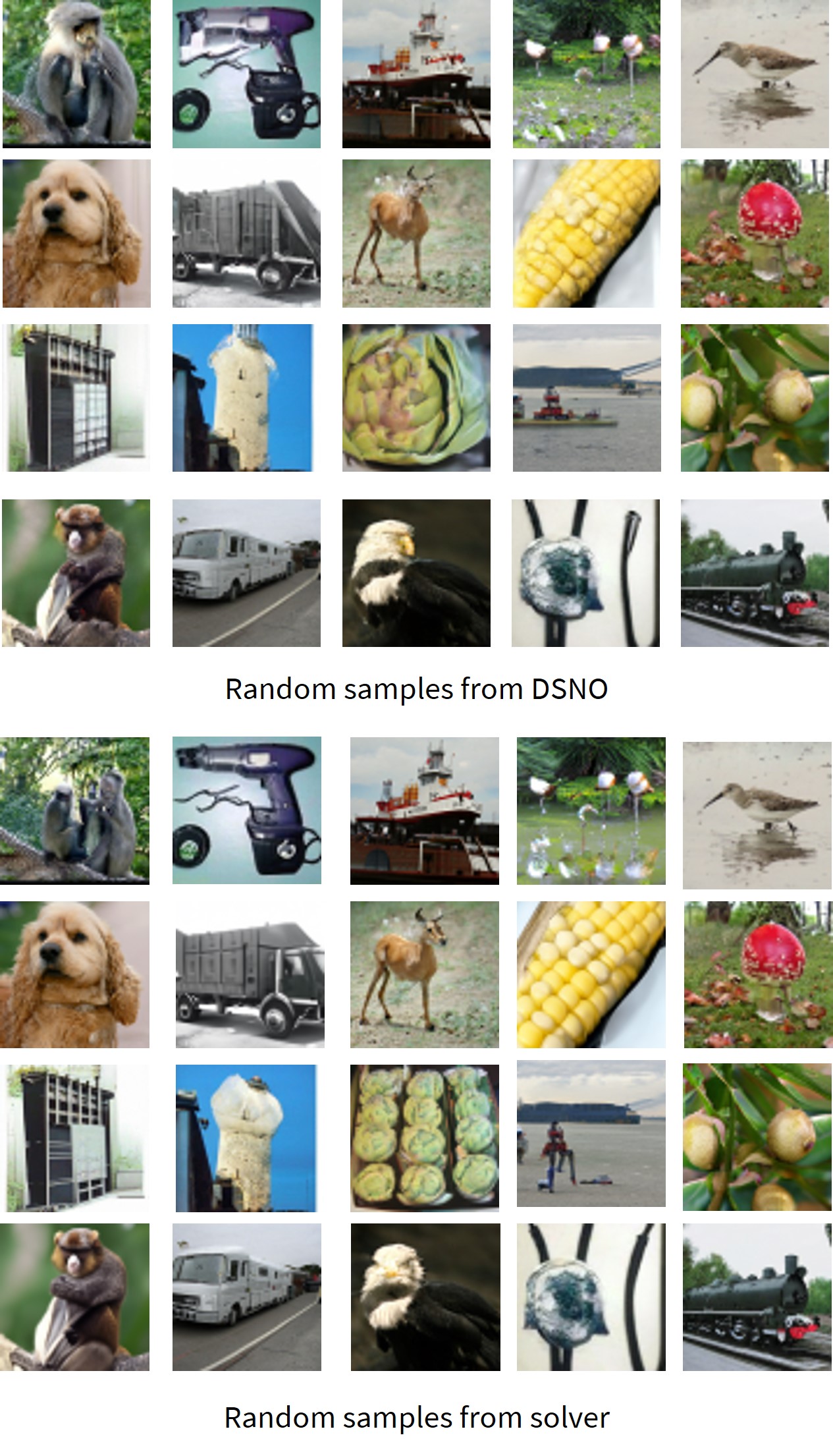}
        \caption{Upper panel: random samples generated by DSNO. Bottom panel: generated by the solver. }
        \label{fig:demo_imagenet_recon}
\end{figure}

\vspace{-0.1in}
\paragraph{Trajectory prediction and reconstruction.}
Figure~\ref{fig:demo_imagenet_traj} compares the trajectories predicted by DSNO and the original ODE solver, respectively, for the fixed random seed with a temporal resolution 4. We see that the DSNO predicted trajectory highly matches the groundth-truth ODE trajectory, which demonstrates the effectiveness of DSNO with parallel decoding. Besides, Figure~\ref{fig:demo_imagenet_recon} shows the random samples from DSNO and the original pre-trained diffusion model with the same random seed. It is clear that the mapping from Gaussian noise to the output image is well-preserved.

\begin{table}[ht]
\caption{Impact of temporal convolution. We compare the performance of architectures with and without temporal convolution blocks while keeping all other settings the same. }
\label{tb:cifar_ablation_temporal_conv}
\begin{center}
\begin{small}
\begin{tabular}{lccc}
\toprule
 Training steps & U-Net &  U-Net + Temporal Conv \\
\midrule
300k  & 8.09 & 4.23 \\
400k & 7.85 & 4.12 \\ 
\bottomrule
\end{tabular}
\end{small}
\end{center}
\vskip -0.1in
\end{table}

\begin{table}[ht]
\caption{Ablation study on the choice of training loss weighting and time discretization scheme. The temporal resolution is fixed to 4 in this group of experiments. }
\label{tb:cifar_ablation_loss-weight}
\begin{center}
\begin{small}
\begin{tabular}{lcc}
\toprule
 Loss weighting & Uniform & $\mathrm{SNR}^{0.5}$ \\
FID  & 4.56 & 4.21 \\
\midrule
time discretization scheme & Uniform & Quadratic \\
FID & 4.33 & 4.21 \\
\bottomrule
\end{tabular}
\end{small}
\end{center}
\vskip -0.1in
\end{table}

\begin{table}[ht]
\caption{Ablation study on the choice of the temporal resolution.}
\label{tb:cifar_ablation_resolution}
\begin{center}
\begin{small}
\begin{tabular}{lccc}
\toprule
Temporal resolution & 2 & 4 & 8 \\
\midrule
FID  & 5.01 & 4.21  & 3.98\\
\bottomrule
\end{tabular}
\end{small}
\end{center}
\vskip -0.1in
\end{table}

\subsection{Ablation study}
In this section, we study the effect of different model choices, including the temporal convolution blocks, loss weighting function, temporal resolution, time discretization scheme, and loss function, by performing ablation studies on CIFAR-10. Without stated explicitly, we use batch size 128 and $\ell^1$ norm for the loss function. 
\vspace{-0.1in}
\paragraph{Temporal convolution block.} We first investigate the impact of temporal convolution by comparing the performance of architectures with and without temporal convolution blocks. All the other settings are kept the same such as temporal resolution 4, quadratic time discretization scheme, the square root of the SNR weighting function, and batch size 256. As reported in Table~\ref{tb:cifar_ablation_temporal_conv}, the temporal convolution design is crucial to DSNO's performance as its kernel integration operator nature is a better model inductive bias to model the trajectory in time.

\vspace{-0.1in}
\paragraph{Loss weighting.}  The loss weighting function used in the training objective of Diffusion models\cite{ho2020denoising, song2020score} typically distributes more weights to the small times, which is important to training diffusion models. We also adopt such a weighting function since it is generally harder to control the error at small times. We observe that such loss weighting function benefits the training of DSNO. As reported in Table~\ref{tb:cifar_ablation_loss-weight}, using the square root of the SNR weighting function slightly improves the FID by 0.35. 
\vspace{-0.1in}
\paragraph{Time discretization scheme.} How to discretize the temporal domain is important to the performance of the numerical solvers. Some small changes to the time discretization scheme could lead to very different sample qualities as shown in~\cite{karras2022elucidating,zhang2022fast}. DSNO also needs to choose a way to discretize the temporal domain. Here we consider the two most common choices of time discretization schemes in the literature: uniform time step and quadratic time step. As shown in Table~\ref{tb:cifar_ablation_loss-weight}, the quadratic time step is slightly better than the uniform time step by 0.12, showing that DSNO is not sensitive to the different time discretization schemes used in the existing solvers and can work nicely with different solvers. 
\vspace{-0.1in}
\paragraph{Temporal resolution.} We study the effect of temporal resolution (i.e., the discretization steps $M$), given the square root of SNR weighting function and the quadratic time discretization scheme. 
As reported in Table~\ref{tb:cifar_ablation_resolution}, the FID improves as we increase the temporal resolution. Since the higher temporal resolution introduces more supervision into the training, it is reasonable to expect better FID scores. However, higher resolution also results in higher computation costs. Since increasing the resolution from 4 to 8 only provides a marginal benefit (due to the compact spectrum we observed in Appendix~\ref{sec:spectrum}), one may use temporal resolution 4 for better efficiency.   

\begin{table}[ht]
\caption{Ablation study on the choice of the loss function. For LPIPS, we use the original VGG-based version without any calibration~\cite{zhang2018unreasonable}.}
\label{tb:cifar_ablation_loss}
\begin{center}
\begin{small}
\begin{tabular}{lcc}
\toprule
Loss function & $\ell^1$ & LPIPS \\
\midrule
FID   & 4.12  & 3.78 \\
\bottomrule
\end{tabular}
\end{small}
\end{center}
\vskip -0.1in
\end{table}
\paragraph{Loss function.} We only vary the loss function but keep the other settings the same, including a batch size of 256, a temporal resolution of 4, and a quadratic discretization scheme.  As shown in Table~\ref{tb:cifar_ablation_loss}, using the original VGG-based LPIPS loss~\cite{zhang2018unreasonable} instead of the standard $\ell^1$ loss leads to a further improvement in the FID score. 


\section{Related work}

\paragraph{ODE-based sampling.}
ODE-based samplers are much more widely used in practice \cite{stablediffusion} than SDE-based methods because they can take large time steps by leveraging some useful structures of the underlying ODE such as semi-linear structure and the form of exponentially weighted integral \cite{lu2022dpm, zhang2022fast}. Existing works \cite{song2021denoising, bao2021analytic, zhang2022fast,dockhorn2022genie} have greatly reduced the number of discretization steps to 10-50 in time while keeping the approximation error small to generate high-quality samples. The exponentially weighted integral structure of the solution trajectory revealed by prior works also inspired our design of the temporal convolution block. 
\vspace{-0.1in}
\paragraph{Operator learning for solving PDEs.}
Neural operators are deep learning models that are designed for mappings between function spaces, i.e., continuous functions~\cite{li2020neural,kovachki2021universal}. They are widely deployed as the de facto deep learning models in scientific computing when dealing with partial differential equations (PDE).
Among these methods, Fourier neural operator stands out and is one of the most efficient machine learning methods for scientific computing problems involving PDE
\cite{yang2021seismic,wen2022accelerating}. It is shown to possess the crucial discretization invariance and universal approximation properties of universal operators~\citep{kovachki2021universal,kovachki2021neural}, which motivates our design of the temporal convolution block in our method.

\vspace{-0.1in}
\paragraph{Training-based sampling.} Training-based methods typically train a neural network surrogate to replace some parts of the numerical solver or even the whole solver. This category includes various methods from diverse perspectives such as knowledge distillation \cite{luhman2021knowledge, salimans2021progressive}, learning the noise schedule \cite{lam2021bilateral, watson2021learning}, learning the reverse covariance \cite{bao2022estimating}, which require extra training. Training-based methods usually work in the few-step regime with less than 10 steps. Direct \citet{luhman2021knowledge} is the first work to get descent sample quality on CIFAR10 with one model evaluation but it suffers from overfitting and its sampling quality drops dramatically compared to the original sampling methods of diffusion models. The current SOTA progressive distillation \cite{salimans2021progressive} reduces the number of steps down to 4-8 without losing much sample quality. However, it has the same issue as knowledge distillation in the limit of one function evaluation. Some other methods~\cite{xiao2021tackling,vahdat2021score, zheng2022truncated} combine diffusion models with other generative models such as GAN and VAE to enable fast sampling. 

\section{Conclusion and discussion}
In this paper, we propose \textit{diffusion model sampling with neural operator} (DSNO) that maps the initial condition, i.e., Gaussian distribution, to the continuous-time solution trajectory of the reverse diffusion process. To better model the temporal correlations along the trajectory, we introduce temporal convolution layers into the given diffusion model backbone.  Experiments show that our method achieves the SOTA FID score of 3.78 for CIFAR-10 and 7.83 for ImageNet-64 with only one model evaluation. 
Our method is a big step toward real-time sampling of diffusion models, which can potentially benefit many time-sensitive applications of diffusion models.

\section*{Acknowledgements}
We would like to thank the reviewers and the area chair for their constructive comments. Anima Anandkumar is supported in part by Bren professorship. This work was done partly during Hongkai Zheng's internship at NVIDIA.




\nocite{langley00}

\bibliography{icml_ref}
\bibliographystyle{icml2023}

\newpage
\appendix
\onecolumn

\section{Appendix}

\subsection{Energy spectrum}
\label{sec:spectrum}
\begin{figure}[htbp]
    \centering
    \includegraphics[width=0.5\columnwidth]{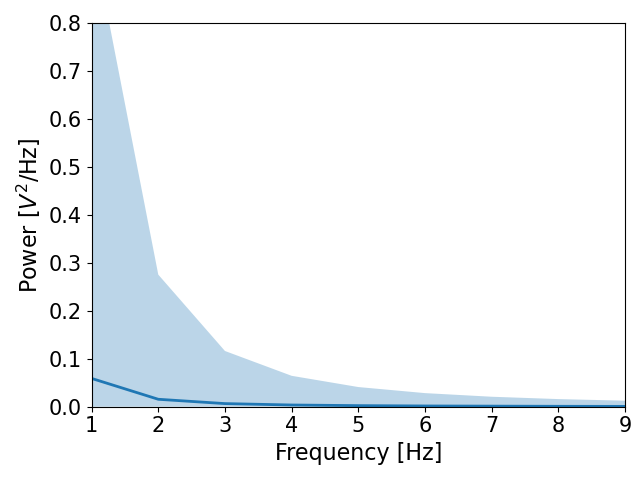}
    \caption{Power spectrum of the ODE trajectories sampled from "DDPM++ cont. (VP)" model trained by \cite{song2020score} on CIFAR10. The mean is computed over all pixel locations and channels of randomly generated trajectories. Most power concentrates in the $\leq 5$ Hz regime. The shaded region represents the maximum and minimum power.  }
    \label{fig:spectrum}
\end{figure}

The discrete-time Fourier transform of the signal $x(t)$ with period $T$ is given by 
\begin{equation}
    X_j = \sum_{i=1}^{N} x(t_i) \exp\left(-\frac{2\pi}{T} ji t_i\right), 
\end{equation}
where $t_i = \frac{iT}{N}$. $\frac{j}{T}$ is the frequency. $j$ is called the frequency mode. Let $\Delta = \frac{1}{N}$ be the time step. The spectrum is defined as the product of the Fourier transform of $x$ with its conjugate:
\begin{equation}
    S_{j} = \frac{2\Delta^2}{T} X_j X_j^{*},
\end{equation}
where $X_j^{*}$ is the complex conjugate. In practice, the statistics are computed over all pixel locations and channels of randomly generated trajectories. $T=1$ and the sampling frequency is 1000 Hz to avoid aliasing. Figure~\ref{fig:spectrum} visualizes the energy spectrum of ODE trajectories sampled from the diffusion model "DDPM++ cont. (VP)" trained by \cite{song2020score} on CIFAR10. We observe that most power concentrates in the regime where the frequency mode is less than 5. 

\subsection{Background: neural operators}
\label{sec:neural_operator}
Let $\mathcal{A}$ and $\mathcal{U}$ be two Banach spaces and $G: \mathcal{A} \rightarrow \mathcal{U}$ be a non-linear map. 
Suppose we have a finite collection of data $\{a_i, u_i\}_{i=1}^{N}$ where $a_i\sim \mu$ are i.i.d. samples from the distribution $\mu$ supported on $\mathcal{A}$ and $u_i=G(a_i)$.
Neural operators aim to learn $G_{\phi}$ parameterized by $\phi$ to approximate $G$ from the observed data by minimizing the empirical risk given by
\begin{equation}
    \min_{\phi} \mathbb{E}_{a\sim \mu} \|G(a) - G_{\phi}(a)\|_{\mathcal{U}} \approx \min_{\phi} \frac{1}{N} \sum_{i=1}^{N} \|u_i - G_{\phi}(a_i)\|_\mathcal{U}.
\end{equation}

The architecture of neural operators is constructed as a stack of kernel integration layers where the kernel function is parameterized by learnable weights. This architecture utilizes the convolution theorem on abelian groups. Among different neural operator architectures, Fourier neural operator~\cite{li2020fourier} stands out and is one of the most efficient machine learning methods for scientific computing problems involving PDE
\cite{yang2021seismic,wen2022accelerating}. It is shown to possess the crucial discretization invariance and universal approximation properties of universal operators~\citep{kovachki2021universal,kovachki2021neural}.

\subsection{Extended set of generated samples}
\begin{figure}[htbp]
    \centering
    \includegraphics[width=0.6\textwidth]{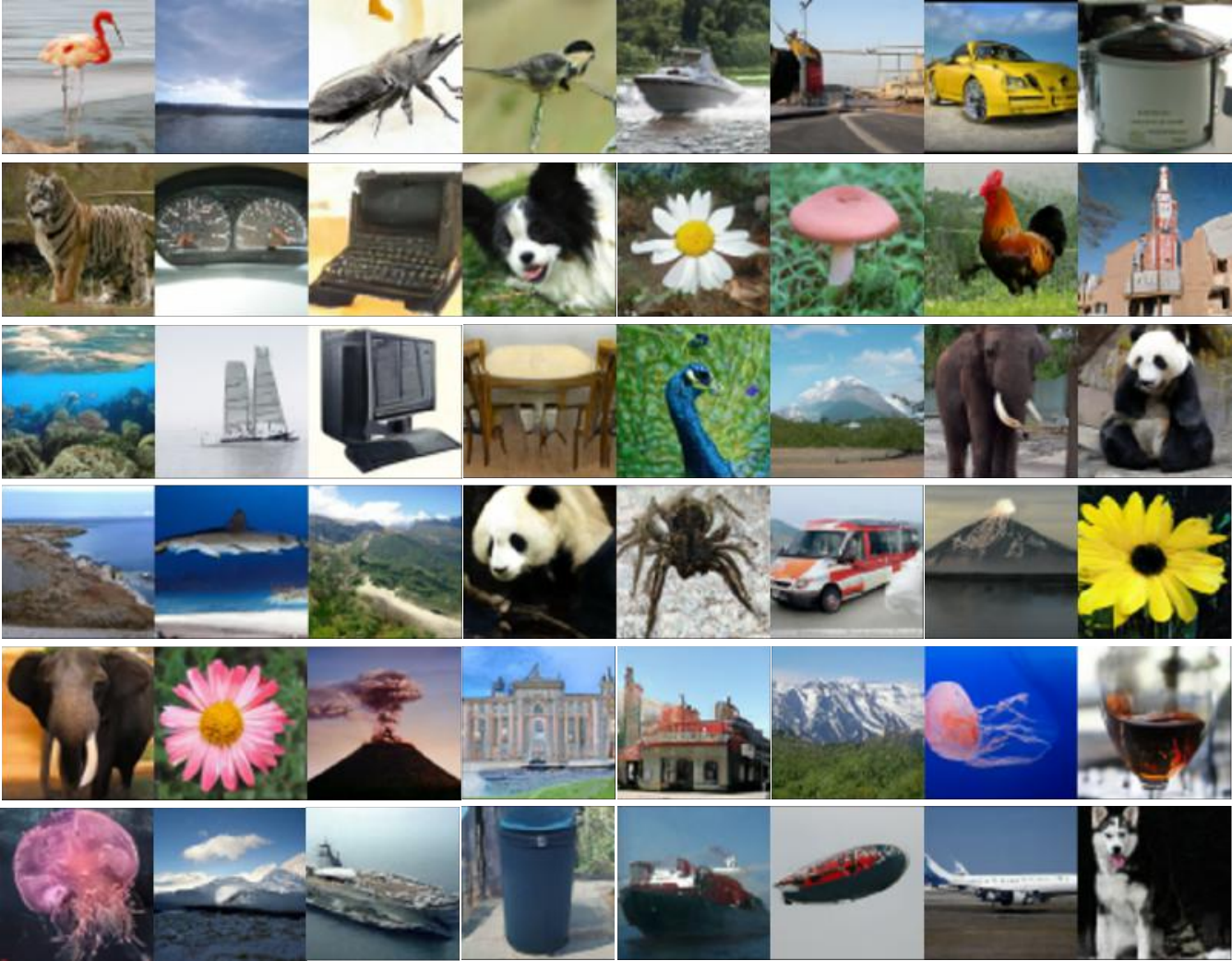}
    \caption{Random samples generated from DSNO on ImageNet-64.}
    \label{fig:images_set1}
\end{figure}

We provide an extended set of randomly generated samples from our ImageNet-64 model in Figure~\ref{fig:images_set1}. 

\subsection{Generalization to different resolution}

\begin{figure*}[ht]
        \centering
        \includegraphics[width=0.75\textwidth]{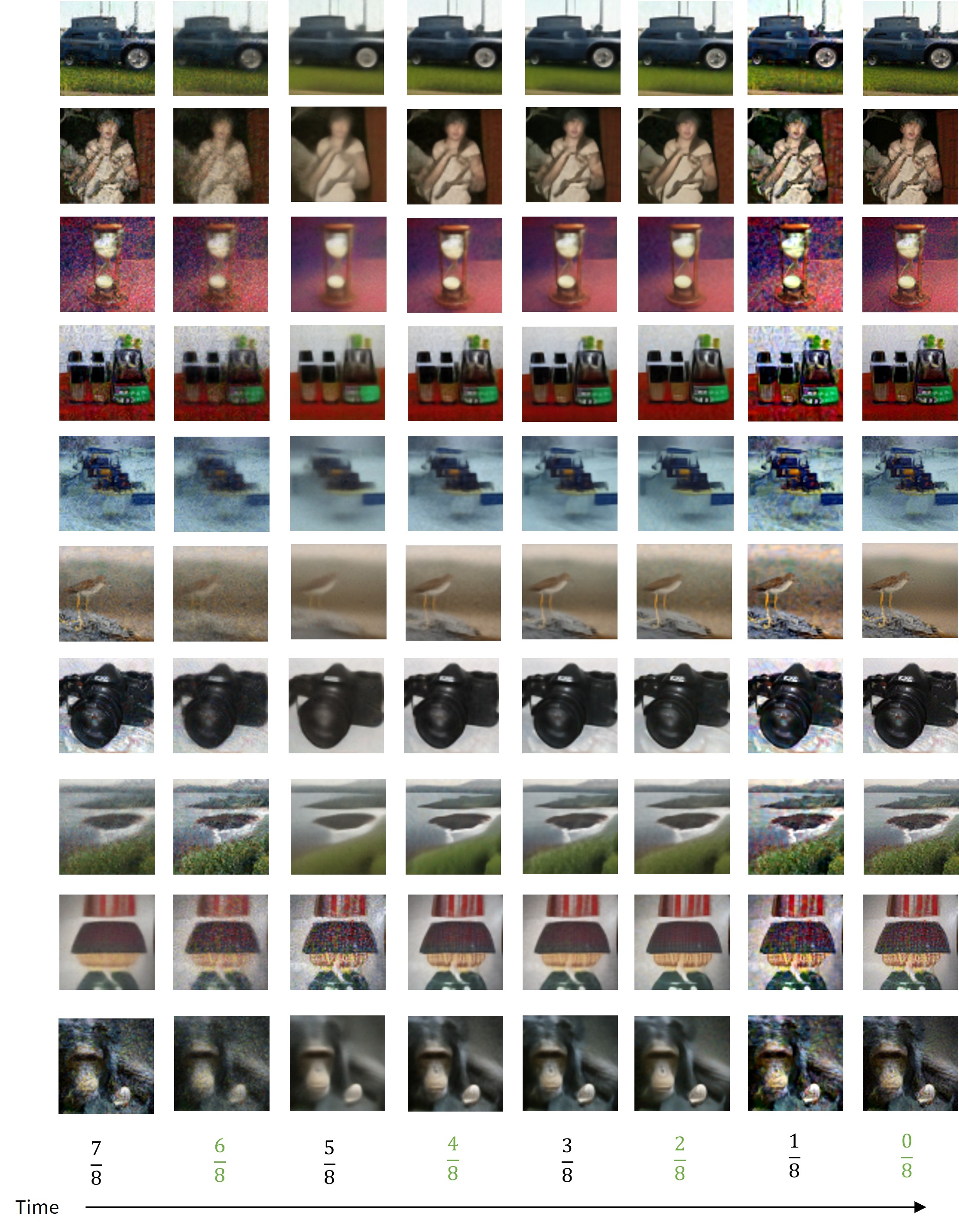}
        \caption{The predicted trajectory of DSNO with a temporal resolution of 8 on ImageNet-64. We train the DSNO with a temporal resolution of 4 and then use it to predict the trajectory with a temporal resolution of 8. Time locations marked green are the points that DSNO is trained with. Black time locations are the points that DSNO never saw in the training set. }
        \label{app_fig:demo_imagenet}
\end{figure*}

Figure~\ref{app_fig:demo_imagenet} visualizes the predicted trajectory of DSNO in temporal resolution 8 on ImageNet-64 while it is trained on temporal resolution 4. Although the resulting trajectories do not look perfectly smooth, it still demonstrates the generalization ability of DSNO to unseen time resolutions.

\subsection{Further discussion}
\paragraph{Future work}
There are several directions we leave as future work. First, guided sampling of diffusion models is widely used in various applications but accelerating guided sampling is also more challenging~\cite{meng2022distillation}. How to adapt DSNO for sampling Guided diffusion model will be an interesting next step. 
Second, the temporally continuous output of DSNO provides another level of flexibility compared to distillation-based methods and is readily available for applications such as DiffPure~\cite{nie2022DiffPure} that require fast forward/backward sampling from diffusion models at various temporal locations. DSNO could potentially reduce the inference time in those applications.
We leave the exploration of those applications to future work. Last but not least, transformer-based architectures have shown their promising capacity for diffusion models~\cite{peebles2022scalable, bao2023all} in high-resolution image generation. It is natural to integrate our temporal convolution layers into these diffusion transformers as the temporal blocks operate solely on the temporal dimension regardless of how the pixel space is modeled. The resulting new architecture could also potentially serve as a new architecture design for other problems where the dynamics are continuous in time. 

\paragraph{Reducing data collection cost with advanced solvers.} While we primarily use DDIM solver to collect data for fair comparison in this paper, it is worth noting that advanced numerical solvers like DPM solver\cite{lu2022dpm} can approximate the solution operator with less computation cost, which will greatly speed up our training data generation process. Our final implementation includes examples of using DPM solvers in our GitHub repository \href{https://github.com/devzhk/DSNO-pytorch}{https://github.com/devzhk/DSNO-pytorch}.

\end{document}